\pdfoutput=1
\documentclass[11pt]{article}

\usepackage[]{acl}

\usepackage{times}
\usepackage{latexsym}

\usepackage[T1]{fontenc}
\usepackage[utf8]{inputenc}

\usepackage{microtype}

\def\ie{{\em i.e.,~}}

\usepackage{subcaption}
\usepackage{amssymb}
\usepackage{graphicx}
\usepackage{comment}
\usepackage{multirow}
\usepackage{amsmath, amsmath}
\DeclareMathOperator*{\argmax}{arg\,max}

\usepackage{xcolor}
\newcommand{\dejiao}[1]{{\color{black}{#1}}}
\newcommand{\needmod}[1]{{\color{black}{#1}}}
\usepackage{todonotes}

\title{Virtual Augmentation Supported Contrastive Learning of Sentence Representations}

\author{Dejiao Zhang\thanks{$^*$ This work is accepted to Findings of ACL 2022. Our code and checkpoints will soon be released at \url{https://github.com/amazon-research/sentence-representations}. Correspondence to Dejiao Zhang <dejiaoz@amazon.com>.} \quad Wei Xiao \quad  Henghui Zhu  \quad 
\textbf{Xiaofei Ma} \quad 
\textbf{Andrew O. Arnold} \\
  AWS AI Labs, New York
  }

\begin{document}
\maketitle
\begin{abstract}
Despite profound successes, contrastive representation learning relies on carefully designed data augmentations using domain specific knowledge. This challenge is magnified in natural language processing where no general rules exist for data augmentation due to the discrete nature of natural language. We tackle this challenge by presenting a Virtual augmentation Supported Contrastive Learning of sentence representations (VaSCL). Originating from the interpretation that data augmentation essentially constructs the neighborhoods of each training instance, we in turn utilize the neighborhood to generate effective data augmentations. Leveraging the large training batch size of contrastive learning, we approximate the neighborhood of an instance via its K-nearest in-batch neighbors in the representation space. We then define an instance discrimination task regarding this neighborhood and generate the virtual augmentation in an adversarial training manner. We access the performance of VaSCL on a wide range of downstream tasks, and set a new state-of-the-art for unsupervised sentence representation learning. 
% \footnote{The code will be released at \url{https://github.com/amazon-research/sentence-representations}.} 
\end{abstract}

\section{Introduction}
Universal sentence representation learning has been a long-standing problem in Natural Language Processing (NLP). Leveraging the distributed word representations \citep{bengio2003neural,mikolov2013distributed,collobert2011natural,pennington2014glove} as the base features to produce sentence representations is a common strategy in the early stage. However, these approaches are tailored to different target tasks, and thereby yielding less generic sentence representations \citep{yessenalina2011compositional,socher2013recursive, kalchbrenner2014convolutional,cho2014properties}. 

% In more recent literature, the progress is largely driven by supervised learning on the Natural Language Inference (NLI) datasets \citep{conneau2017supervised,cer2018universal,reimers2019sentence,gao2021simcse,zhang2021pairwise}. Despite promising progress,  the high cost of collecting annotations limits its generalizability, and moreover the transfer learning performance degrade when the target domain differs significantly
% from the NLI dataset used for training \citep{zhang2020unsupervised}.    
This issue has motivated more research efforts on designing generic sentence-level learning objectives or tasks. Among them, supervised learning on the Natural Language Inference (NLI) datasets \citep{bowman2015large, williams2017broad,wang2018glue} has established benchmark transfer learning performance on various downstream tasks \citep{conneau2017supervised,cer2018universal,reimers2019sentence,zhang2021pairwise}. Despite promising progress,  the high cost of collecting annotations precludes its wide applicability, especially in the scenario where the target domain has scarce annotations but differs significantly from the NLI dataset \citep{zhang2020unsupervised}.

On the other hand, unsupervised learning of sentence representations has seen a resurgence of interest with the recent successes in self-supervised contrastive learning. 
These approaches rely on two main components,  data augmentation and \dejiao{an instance-level contrastive loss}.
%contrastive loss that aims to separate each instance and its augmentations apart from the others. 
The popular contrastive learning objectives \citet{chen2020simple,he2020momentum} and their variants thereof, have empirically shown their effectiveness in NLP. However, the discrete nature of text makes it challenging to establish universal rules for \dejiao{effective text augmentation generation}.

Various contrastive learning based approaches have been proposed for learning sentence representations, with the main difference lies in how the augmentations are generated \citep{fang2020cert,giorgi2020declutr,wu2020clear,meng2021coco,yan2021consert,kim2021self,gao2021simcse}. \dejiao{Somewhat surprisingly, a recent work \citep{gao2021simcse} empirically shows that Dropout \citep{srivastava2014dropout}, \ie \textit{augmentations obtained by feeding the same instance to the encoder twice},  outperforms common augmentation strategies operating on the text directly, including cropping, word deletion, or synonym replacement.}  On the other side of the coin, this observation again validates the inherent difficulty of \dejiao{attaining effective data augmentations in NLP.} 

%instead of relying on explicit operations on the discrete text, 
In this paper, we tackle the challenge by presenting a neighborhood guided virtual augmentation strategy to support contrastive learning. In a nutshell, data augmentation essentially constructs the neighborhoods of each instance, with the semantic content being preserved. We take this interpretation in the opposite direction by leveraging the neighborhood of an instance to \dejiao{guide augmentation generation}. Benefiting from the large training batch of contrastive learning, we approximate the neighborhood of an instance via its K-nearest in-batch neighbors. We then define an instance discrimination task within this neighborhood and generate the virtual augmentation in an adversarial training manner. We evaluate our model on a wide range of downstream tasks, and show that our model consistently outperform the previous state-of-the-art results by a large margin. We run in-depth analysis and show that our VaSCL model leads to a more dispersed representation space with the data semantics at different granularities being better captured. 

\begin{figure*}
    \centering
    \includegraphics[scale=0.46]{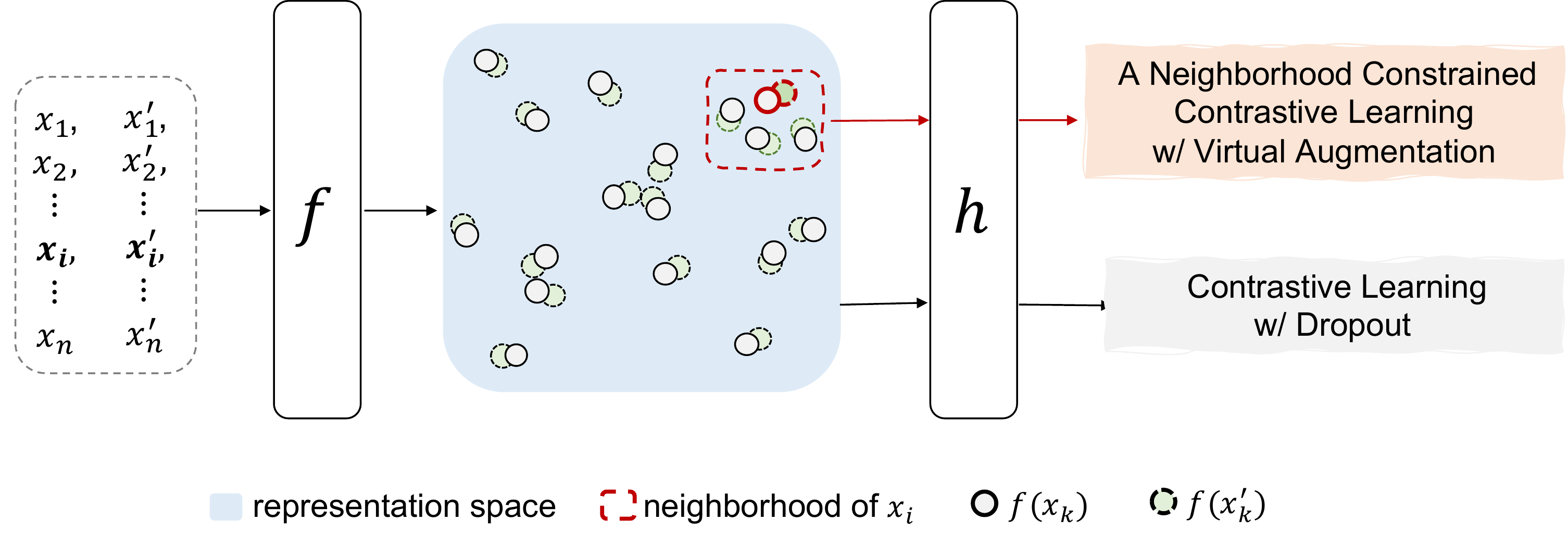}
    \caption{Illustration of VaSCL. For each instance $x_i$ in a randomly sampled batch, we optimize (i) an instance-wise contrastive loss with the dropout induced augmentation obtained by forwarding the same instance twice, \ie $x_i$ and $x_{i'}$ denote the same text example; and (2) a neighborhood constrained instance discrimination loss with the virtual augmentation proposed in Section \ref{subsec:NCL_virtual_model}. }
    \label{fig:model_illustration}
\end{figure*}

\section{Related Work}
\paragraph{Universal Sentence Representation Learning}{Arguably, the simplest and most common approach for attaining sentence representations are bag-of-words \citep{harris1954distributional} and variants thereof. However, bag-of-words suffers from data sparsity and lack of sensibility to word semantics. In the past two decades, the distributed word representations \citep{bengio2003neural,mikolov2013distributed,collobert2011natural,pennington2014glove} 
have become the more effective base features for producing sentence representations.  The downside is \dejiao{that} these approaches are tailored to \dejiao{the} target tasks \citep{yessenalina2011compositional,socher2013recursive, kalchbrenner2014convolutional,cho2014properties}, and thereby the resulting sentence representations attain limited transfer learning performance. 

More recent efforts focus on directly designing sentence-level learning objectives or tasks.  On the supervised learning regime, \citet{conneau2017supervised,cer2018universal} empirically show the effectiveness of leveraging the NLI task \citep{bowman2015large, williams2017broad} to promote generic sentence representations. The task involves classifying each sentence pair into one of three categories: entailment, contradiction, or neutral. \citet{reimers-2019-sentence-bert} further \dejiao{bolster} the performance by using the pre-trained transformer \citep{devlin2018bert,liu2019roberta} as backbone. On the other end of spectrum, \citet{hill-etal-2016-learning,bowman2015generating} propose \dejiao{ using the denoising or variational autoencoders for sentence representation learning.}   \citet{kiros2015skip,hill-etal-2016-learning} extend the distributional hypothesis to the sentence-level, and train an encoder-decoder to construct surrounding context for each sentence.  Alternatively, \citet{logeswaran2018efficient} present a model that learns to discriminate the target context sentences from all contrastive ones.}

\paragraph{Contrastive Learning}
Contrastive learning has been the pinnacle of recent successes in sentence representation learning. \citet{gao2021simcse,zhang2021pairwise} substantially advance the previous state-of-the-art results by leveraging the entailment sentences in NLI as positive pairs for optimizing the properly designed contrastive loss functions. Nevertheless, we focus on unsupervised contrastive learning and form the positive pairs via data augmentation, since such methods are more cost-effective and applicable across different domains and languages. 
\dejiao{Along this line, several approaches have been proposed recently, where the augmentations are obtained via dropout \citep{yan2021consert,gao2021simcse}, back-translation \citep{fang2020cert}, surrounding context sampling \citep{logeswaran2018efficient,giorgi2020declutr}, or perturbations conducted at different semantic-level \citep{wu2020clear,yan2021consert,meng2021coco}.
% or intermediate representations of the transformer \citep{kim2021self}.
}

\paragraph{Consistency Regularization}
% Our work is also closely related to both consistency regularization and data augmentation.
% The two directions themselves are highly correlated 
% as consistency regularization forces the model
% output to remain unchanged under plausible variations that are often induced via data augmentations, while different augmentations of a instance should be consistent with the underlying semantics, in order to be valid.  
Our work is also closely related to consistency regularization, which is often used to promote better performance by regularizing the model
output to remain unchanged under \dejiao{plausible input variations that are often induced via data augmentations.}
\citet{bachman2014learning,sajjadi2016regularization,samuli2017temporal,tarvainen2017mean} show randomized data augmentation such as dropout, cropping, rotation, and flipping yield effective regularization. \citet{berthelot2019mixmatch,Berthelot2020ReMixMatch:,verma2019interpolation} bolster the performance by applying Mixup \citep{zhang2017mixup} and its variants on top of stochastic data augmentations. 
% However, data augmentation has long been a challenge in NLP as there is no general rules for effective text transformations with the semantic content being preserved.
% An alternative come to light when one considers the violation of consistency regularization can in turn be used to find the virtual augmentation that the model is most sensitive to. In this paper, we utilize the consistency regularization to promote strong augmentation for an instance, while leveraging its approximated neighborhood to generate valid augmentations that share the same semantic content. 
\needmod{However, data augmentation has long been a challenge in NLP as there is no general rules for effective text transformations.
An alternative come to light when considering the violation of consistency regularization can in turn be used to find the perturbation that the model is most sensitive to. In this paper, we utilize the consistency regularization to promote informative virtual augmentation for an instance in the representation space, while leveraging its approximated neighborhood to regularize the  augmentation sharing the similar semantic content as its original instance.} 

% Another relevant line of work is consistency regularization, which is often used to promote better generalization by regularizing the model
% output remain unchanged against plausible input variations (augmentations).  \citet{samuli2017temporal,sajjadi2016regularization}   shows randomized data augmentation such as cropping, rotation, or flipping yield effective regularization. \citet{berthelot2019mixmatch,verma2019interpolation} bolster the performance by applying Mixup \citep{zhang2017mixup} on top of stochastic data augmentations. Instead of relying on explicit data augmentations, \citet{bachman2014learning,samuli2017temporal,tarvainen2017mean} show empirical successes by simply using Dropout \citep{srivastava2014dropout} as pseudo data augmentations.  \citet{miyato2016adversarial,miyato2018virtual} strengthen the augmentations by approximating the perturbation direction in the input space that the model is most sensitive to. 

\section{Method}
\subsection{Preliminaries}
Self-supervised contrastive learning often aims to solve the instance discrimination task. 
In our scenario, let $f$ denote the transformer encoder that maps the $i^{\text{th}}$ input sentence $\mathbf{x}_i$ to its representation vector $\mathbf{e}_i = f(\mathbf{x}_i)$\footnote{By an abuse of notation, we assume $f$ outputs either the pre-defined sentence representation (a.k.a. [CLS] embedding, \citep{devlin2018bert}), or the mean/max pooling of all tokens' embeddings of that sentence.}. Further let $h$ be the contrastive learning head and $\mathbf{z}_i = h(f(\mathbf{x}_i))$ denote the final output for $\mathbf{x}_i$.  Let $\mathcal{B} = \left\{i, {i'}\right\}_{i=1}^{M}$ denote the indices of a randomly sampled batch of paired examples, where $\mathbf{x}_i, \mathbf{x}_{i'}$ are two independent variations of the $i^{\text{th}}$ instance. A popular loss function \citep{chen2020simple}  for contrastive learning is defined as follows, 
\begin{align}
\label{eq:instance_disc_orig}
    &\ell_{\mathcal{B}}(\mathbf{z}_{i}, \mathbf{z}_{i'}) = \\
    &\quad -\log \frac{e^{\mathbf{sim}(\mathbf{z}_{i}, \mathbf{z}_{i'})/\tau}}{e^{\mathbf{sim}(\mathbf{z}_{i}, \mathbf{z}_{i'})/\tau} + \sum_{j \in \mathcal{B}\setminus (i,{i'})} e^{\mathbf{sim}(\mathbf{z}_{i}, \mathbf{z}_{j})/\tau}}\;, \nonumber 
\end{align}
where $\tau$ is the temperature hyper-parameter and $\mathbf{sim}(\cdot)$ denotes the cosine similarity, \ie $\mathbf{sim}(\cdot) = \mathbf{z}_i^T\mathbf{z}_{i'} / \|\mathbf{z}_i\|_2 \|\mathbf{z}_{i'}\|_2$.  Similarly, $\ell_{\mathcal{B}}(\mathbf{z}_{i'}, \mathbf{z}_{i})$ is defined by exchanging the roles of $\mathbf{z}_i$ and $\mathbf{z}_{i'}$ in the above equation.
Intuitively, Equation (\ref{eq:instance_disc_orig}) defines the
log-likelihood of classifying 
the $i^{th}$ instance as its positive $i'$
% $\mathbf{x}_{i'}$ 
among all $2M$--1 candidates within the same batch $\mathcal{B}$. Therefore, minimizing the above log-loss guides the encoder to map each positive pair close in the representation space, and negative pairs further apart.

\paragraph{Dropout based contrastive learning} As Equation (\ref{eq:instance_disc_orig}) implies, the success of contrastive learning relies on effective positive pairs construction. 
% In computer vision, positive pairs generated by heavy data augmentations have shown to be essential to the top-performing contrastive learning models \citep{chen2020simple,he2020momentum}.
However, it's difficult to generate strong and effective data transformations in NLP due to the discrete nature of natural language. This challenge is further demonstrated in a recent work \citep{gao2021simcse}, which shows that augmentations obtained by Dropout \citep{srivastava2014dropout}, \ie \textit{$\mathbf{z}_i, \mathbf{z}_{i'}$ obtained by forwarding the same instance $\mathbf{x_i}$ twice},  outperforms the common text augmentation strategies such as cropping, word deletion, or synonym replacement.
% On the other hand, a recent work \citep{gao2021simcse} shows that augmentations obtained by Dropout \citep{srivastava2014dropout}, \ie \textit{$\mathbf{z}_i, \mathbf{z}_{i'}$ obtained by forwarding the same instance $\mathbf{x_i}$ twice},  outperforms the common text augmentation strategies such as cropping, word deletion, or synonym replacement. 
Dropout provides a natural way for data augmentation by randomly masking its inputs or the hidden layer nodes. The effectiveness of using Dropout as pseudo data augmentations can be traced back to  \citet{bachman2014learning,samuli2017temporal,tarvainen2017mean}. Nevertheless, the augmentation strength is weak with dropout only, there is room for improvement, which we investigate in the following section.

\subsection{Neighborhood Constrained Contrastive Learning with Virtual Augmentation}
\label{subsec:NCL_virtual_model}
In essence, data augmentation can be interpreted as constructing the neighborhood of a training
instance, with the semantic content being preserved. In this section, we take this interpretation in the opposite direction by first approximating the neighborhood of an instance, upon which we generate the augmentations. 
% To be more specific, we first approximate the neighborhood $\mathcal{N}(i)$ of the $i^{\text{th}}$ instance as its K-nearest neighbors in the representation space, 
To be more specific, let $\bar{\mathcal{B}} = \left\{i\right\}_{i=1}^{M}$ denote the indices of a randomly sampled batch with $M$ examples. We first approximate the neighborhood $\mathcal{N}(i)$ of the $i^{\text{th}}$ instance as its K-nearest neighbors in the representation space, 
\begin{align*}
    \mathcal{N}(i) &= \left\{k: \mathbf{e}_k ~\text{has the top-K similarity with} ~\mathbf{e}_i \right.\\
    & \left. \qquad \text{among all other M-1 instances in}~ \bar{\mathcal{B}} \right\}
\end{align*}

We then define an instance-level contrastive loss regarding the $i^{\text{th}}$ instance and its neighborhood as follows, 
\begin{align}
\label{eq:neighbor_loss_defn}
    &\ell_{\mathcal{N}(i)}(\mathbf{z}_{i}^\delta, \mathbf{z}_{i})= \\
    &\quad -\log \frac{e^{\mathbf{sim}(\mathbf{z}_{i}^\delta, \mathbf{z}_i)/\tau}}{e^{\mathbf{sim}(\mathbf{z}_{i}^\delta, \mathbf{z}_i)/\tau} + \sum_{k\in \mathcal{N}(i)} e^{\mathbf{sim}(\mathbf{z}_{i}^\delta, \mathbf{z}_k)/\tau}}\;. \nonumber 
\end{align}
In the above equation, $\mathbf{z}_{i}^\delta = h(\mathbf{e}_{i}^\delta)$ denotes the output of the contrastive learning head with the perturbed representation $\mathbf{e}_{i}^\delta = \mathbf{e}_i + \delta_i$ as the input. Here, the initial perturbation 
$\delta_i$ is chosen as isotropic\footnote{$\delta_i$ is sampled from a multivariate normal distribution with the covariance matrix being a scaled identity matrix.} Gaussian noise. As it implies, Equation (\ref{eq:neighbor_loss_defn}) shows the negative log-likelihood of classifying the perturbed $i^{\text{th}}$ instance as itself rather than its neighbors.  Then the augmentation of the $i^{\text{th}}$ instance is retained by identifying the optimal perturbation that maximally disturbs its instance-level identity within this neighborhood. That is,
\begin{align}
\label{eq:optimal_delta}
    &\delta_i^\ast = \argmax_{\|\delta_i\|_2 \leq \Delta}~ \ell_{\mathcal{N}(i)}(\mathbf{z}_{i}^\delta, \mathbf{z}_{i})\;, \\
    &\mathbf{e}_{i^\ast} = \mathbf{e}_i + \delta_i^\ast \;. 
    \label{eq:neighbor_delta}
\end{align}

Denote $\mathcal{N}_{\text{A}}(i)$ as the augmented neighborhood of the $i^{\text{th}}$ instance, \ie  $\mathcal{N}_{\text{A}}(i) = \left\{k, k^\ast\right\}_{k=1}^{K}$ with $\mathbf{e}_{k}$ being the original embedding of the $k^{\text{th}}$ instance and  $\mathbf{e}_{k^\ast}$ denoting the associated augmented representation. Note here $\mathbf{e}_{k^\ast}$ is obtained in the same way defined in Equation (\ref{eq:neighbor_delta}) with respect to its own neighborhood $\mathcal{N}(k)$. 
We then discriminate the $i^{\text{th}}$ instance and its augmentation from the others within its augmented neighborhood $\mathcal{N}_{\text{A}}(i)$ as the following, 
\begin{align}
    \ell_{\mathcal{N}_{\text{A}}(i)} = \ell_{\mathcal{N}_{\text{A}}(i)}(\mathbf{z}_{i}^\ast, \mathbf{z}_{i})  + \ell_{\mathcal{N}_{\text{A}}(i)}(\mathbf{z}_{i}, \mathbf{z}_{i}^\ast) \;.
\end{align}
Here both terms on the right hand side are defined in the same way as Equation (\ref{eq:neighbor_loss_defn}) with respect to the representation of the $i^{th}$ instance and its augmentation ($e_i$, $e_i^\ast$), as well as its neighborhood with the associated optimal augmentations, \ie $\mathcal{N}_{\text{A}}(i)$ defined above.

\paragraph{Putting it all together}{Therefore, for each randomly sampled minibatch $\mathcal{B}$ with $M$ samples, we minimize the following:
\begin{align}
    \mathcal{L}_{\text{VaSCL}} 
    &= \frac{1}{2M} \sum_{i=1}^{M} \left\{ \ell_{\mathcal{\bar{B}}}(\mathbf{z}_{i}, \mathbf{z}_{i'}) + \ell_{\mathcal{\bar{B}}}(\mathbf{z}_{i'}, \mathbf{z}_{i})  
    \right. \nonumber \\
    &\left. \qquad + \ell_{\mathcal{N}_{\text{A}}(i)}(\mathbf{z}_{i}, \mathbf{z}_{i}^\ast) + \ell_{\mathcal{N}_{\text{A}}(i)}(\mathbf{z}_{i}^\ast, \mathbf{z}_{i}) \right\}
    \label{eq:VaSCL_final_loss}
\end{align}
Here $\ell_\mathcal{\bar{B}}(\mathbf{z}_{i}, \mathbf{z}_{i'})$ is defined in the same way as Equation (\ref{eq:instance_disc_orig}), with the only difference being $\mathbf{z}_i, \mathbf{z}_{i'}$ are retained by feeding the $i^{\text{th}}$ instance in $\mathcal{\bar{B}}$ to the encoder twice. Here $\mathbf{z}_{i^\ast}$ is obtained via the virtual augmentation $e_i^\ast$ identified by solving Equations (\ref{eq:optimal_delta})\&(\ref{eq:neighbor_delta}).   Putting together, two instance discrimination tasks are posed for each training example: discriminating each instance and its dropout induced variation from the other in-batch instances; and separating each instance and its virtual augmentation from its top-K nearest neighbors and their virtual augmentations. 
}   

\begin{table*}[htbp]
  \begin{center}
  {\normalsize{
    \begin{tabular}{lcccccccc}
      & \textbf{STS12}& \textbf{STS13}&\textbf{STS14}& \textbf{STS15}&\textbf{STS16}& \textbf{SICK-R}&\textbf{STS-B}& \textbf{Avg.} \\ 
      \cline{2-9} 
    %   GloVe & 51.54&  48.40&  54.21&  57.09&  55.28&  55.42&  51.69&  53.37 \\
      RoBERTa$_{\text{distil}}$ &54.41  &46.85  &56.96  &65.79  &64.22  &61.10  &59.01  &58.33
  \\
    %   {RoBERTa}$_{\text{distil}}$-W &61.68 &69.35 & 66.30&  73.79&  72.06&  62.85&  71.60&  68.23
%  \\
      SimCSE$_{\text{distil}}$ &65.58	&77.42	&70.17	&79.31	&78.45	&67.66	&\textbf{77.98}	&73.79
 \\ 
      \hline 
      \textbf{VaSCL}$_{\text{distil}}$ &\textbf{67.68}  &\textbf{80.61} &\textbf{72.19} &\textbf{80.92} &\textbf{78.59} &\textbf{68.81} &77.32  &\textbf{75.16}

      \\ \\
      RoBERTa$_{\text{base}}$ &53.95  &47.42  &55.87  &64.73  &63.55  &62.94  &58.40  &58.12 \\
      SimCSE$_{\text{base}}$ &68.88	&80.46	&73.54	&80.98	&\textbf{80.68}	&69.54	&\textbf{80.29}	&76.34 
 \\
      \hline 
      \textbf{VaSCL}$_{\text{base}}$ &\textbf{69.08}  &\textbf{81.95} &\textbf{74.64} &\textbf{82.64} &80.57 &\textbf{71.23} &80.23 &\textbf{77.19}
      \\ \\
      
     RoBERTa$_{\text{large}}$ &55.00  &50.14  &54.87  &62.14  &62.99  &58.93  &54.56  &56.95
\\
    %  {RoBERTa}$_{\text{large}}$-W &61.88  &68.67  &65.49  &75.10  &71.92  &64.33  &73.08  &68.64
%  \\
      SimCSE$_{\text{large}}$ &69.83	&81.29	&74.42	&83.77	&79.79	&68.89	&80.66	&76.95 \\
      \hline 
      \textbf{VaSCL}$_{\text{large}}$ &\textbf{74.34}   &\textbf{83.35}   &\textbf{76.79}   &\textbf{84.37}   &\textbf{81.46}   &\textbf{73.23}   &\textbf{82.86}   &\textbf{79.48} \\
    \end{tabular}}}
    \caption{Spearman rank correlation between the cosine similarity of sentence representation pairs and the ground truth similarity scores. 
    % $\diamondsuit$ and $\spadesuit$: results evaluated on the checkpoints provided by \citep{reimers-2019-sentence-bert} and \citep{gao2021simcse}, respectively.
    }
    \label{tab:STS_compare}
  \end{center}
\end{table*}

\section{Experiment}
\label{sec:experiments}
In this section, we mainly evaluate VaSCL against SimCSE \citep{gao2021simcse} which leverages the dropout \citep{srivastava2014dropout} induced noise as data augmentation. We show that VaSCL consistently outperforms SimCSE on various downstream tasks that involve semantic understanding at different granularities.  We carefully study the regularization effects of VaSCL and empirically demonstrate that VaSCL leads to a more dispersed representation space with semantic structure better encoded. Please refer to Appendix \ref{appendix:implementation} for details of the training dataset and our implementations.

\subsection{Evaluation Datasets}
\label{sec:datasets}
% Despite previous work mainly focuses on the semantic textual similarity (a.k.a STS) related tasks, it has been pointed out by several recent work \citep{reimers2016task,wang2021tsdae,zhang2021pairwise} that the performance on STS may not correlate with the other downstream task performances. To provide more comprehensive evaluation, we incorporate two additional downstream tasks, short text clustering and intent classification. 

In addition to the popular semantic textual similarity (a.k.a STS) related tasks, we evaluate two additional downstream tasks, short text clustering and few-shot learning based intent classification. Our motivation is twofold. First, these two tasks provide a new evaluation aspect that complements the pairwise similarity oriented STS evaluation by assessing the high-level categorical semantics encoded in the representations.   Second, two desired challenges are posted as short text clustering requires more effective representations due to the weak signal each text example manifests; and intent classification often suffers from data scarcity since the intents can  vary significantly over different dialogue systems and it is very costly to collect enough intent examples for model training.   
% As shown in Tables \ref{tab:cluster_datastats}\&\ref{tab:intent_datastats}, the collected datasets present the desired diversity in terms of both domains and the class-level statistics. 

\paragraph{Semantic Textual Similarity}
The semantic textual similarity (STS) tasks are the most commonly used benchmark for evaluating sentence representations. STS consists of seven tasks, namely STS 2012-2016 \citep{agirre2012semeval,agirre2013sem,agirre2014semeval,agirre2015semeval,agirre2016semeval}, the STS Benchmark \citep{cer2017semeval}, and the SICK-Relatedness  \citep{marelli2014sick}.  For each sentence pair in these datasets, a fine-grained similarity score ranges from 0 to 5 is provided.

\paragraph{Short Text Clustering} Compared with general text clustering, short text clustering has its own challenge due to lack of signal.  Nevertheless, texts contain only few words grow at unprecedented rates from a wide range of popular resources, including Reddit, Stackoverflow, Twitter, and Instagram. Clustering those texts into groups of similar texts plays a crucial role in many real-world applications such as topic discovery \citep{kim2013discovering}, trend detection \citep{mathioudakis2010twittermonitor}, and recommendation \citep{bouras2017improving}. 
We evaluate on six benchmark datasets for short text clustering. As shown in Table \ref{tab:cluster_datastats}, the datasets present the desired diversities regarding both the cluster sizes and the number of clusters contained in each dataset.

\paragraph{Intent Classification}
Intent classification aims to identify the intents of user utterances, which is a critical component of goal-oriented dialog systems. Attaining high intent classification accuracy is an important step towards solving many downstream tasks such as dialogue state tracking \citep{wu2019transferable,zhang2019find} and dialogue management \citep{gao2018neural,ham2020end}. A
practical challenge is data scarcity because different systems define different sets of intents, and it is costly to obtain enough utterance samples for each intent. Therefore, few-shot learning has attracted much attentions under this scenario, which is also our main focus. We evaluate on four intent classification datasets originating from different domains. We summarize the data statistics in Appendix \ref{appendix:IC_datastats}.

\begin{table*}[htbp]
  \begin{center}
  {\normalsize{
    \begin{tabular}{lccccccc}
      \multirow{2}{*}{} &\textbf{Ag} &\textbf{Search} &{\textbf{Stack}} & {\textbf{Bio-}} &\multirow{2}{*}{\textbf{Tweet}} & {\textbf{Google}}& \multirow{2}{*}{\textbf{Avg}}\\
      
      &\textbf{News} &\textbf{Snippets} &\textbf{Overflow}& \textbf{medical}& & \textbf{News}& \\
      \cline{2-8}
    %   Glove& 82.39  &69.79  &19.54  &28.52  &51.27  &65.29  &52.80 \\
      RoBERTa$_{\text{distil}}$ &59.32  &33.18  &14.16  &24.69  &37.10  &58.05  &37.75
\\
      SimCSE$_{\text{distil}}$&\textbf{73.33}	&60.74	&66.97	&35.69	&50.68	&67.55	&59.16 \\ 
       \hline 
      \textbf{VaSCL}$_{\text{distil}}$ &71.71 &\textbf{62.76}  &\textbf{73.98}  &\textbf{38.82}  &\textbf{51.35} &\textbf{67.66}  &\textbf{61.05} 
      \\ \\
       RoBERTa$_{\text{base}}$ &66.50 &30.83  &15.63  &26.98  &37.80  &58.51  &39.38
 \\
      SimCSE$_{\text{base}}$ &65.53	&55.97	&64.18	&38.12	&49.16	&65.69	&56.44 \\
       \hline 
      \textbf{VaSCL}$_{\text{base}}$ &\textbf{67.46} &\textbf{62.58} &\textbf{73.60}  &\textbf{38.58}  &\textbf{50.98}  &\textbf{66.58}  &\textbf{59.96}
      \\ \\
      
      RoBERTa$_{\text{large}}$ &\textbf{69.35} &\textbf{53.00}  &27.89  &33.25  &46.08  &64.04& 48.93
 \\
      SimCSE$_{\text{large}}$ &62.93	&51.55	&54.11	&35.39	&50.92	&67.86	&53.79 
 \\
      \hline 
      \textbf{VaSCL}$_{\text{large}}$ &57.26  &50.11  &\textbf{76.21}  &\textbf{42.60}  &\textbf{56.10}  &\textbf{69.26}  &\textbf{58.59} \\
    \end{tabular}}}
    \caption{Clustering accuracy reported on six short text clustering datasets.}
    \label{tab:Cluster_Eval}
  \end{center}
\end{table*}

\subsection{Main Results}
\label{sec:compare_with_sota}
\subsubsection{Evaluation Setup}
\textbf{Semantic Textual Similarity.} Same as \citet{reimers-2019-sentence-bert,gao2021simcse}, in Table \ref{tab:STS_compare} we report the Spearman correlation\footnote{Same as  \citet{reimers-2019-sentence-bert,gao2021simcse}, we concatenate all the topics and report the overall Spearman’s correlation.} between the cosine similarity of the sentence representation pairs and the ground truth similarity scores. 
\textbf{Short Text Clustering.}
We evaluate the sentence representations using K-Means \citep{macqueen1967some,lloyd1982least} given its simplicity, and report the clustering accuracy\footnote{The clustering accuracy is computed by using the Hungarian algorithm \citep{munkres1957algorithms}.} averaged over 10 independent runs in Table \ref{tab:Cluster_Eval}. 
\textbf{Intent Classification.}
We freeze the transformer and fine-tune a linear classification layer with the softmax-based cross-entropy loss.   We merge the training and validation sets, from which we sample K training and validation samples per class. We report the mean and standard deviation of the testing classification accuracy evaluated over 5 different splits in Table \ref{tab:ic_fsl_compare}.\footnote{In each setting, we fix the 5 different splits for all models.} We set learning rate to $1e$-04 and batch size to 32. For each task, we train the model with 1000 iterations and evaluate on validation set every 100 iterations. We report the testing accuracy on the checkpoint achieving validation accuracy. 

\begin{table}[htbp]
  \begin{center}
  {\scriptsize{
    % \begin{tabular}{llcccc}
    \begin{tabular}{p{0.1cm}p{1.0cm}p{1.0cm}p{1.2cm}p{1.0cm}p{1.0cm}}
    && \textbf{SNIPS}& \textbf{BANK77} &\textbf{CLINC150} & \textbf{HWU64} \\
      \cline{3-6}
      \multirow{3}{*}{\rotatebox[origin=c]{90}{5-Shot}}
      &RoBERTa &76.71$_{\pm4.84}$   &38.77$_{\pm2.29}$  &55.19$_{\pm1.99}$ &51.52$_{\pm 2}$\\
      &SimCSE&76.94$_{\pm2.53}$  &\textbf{67.48}$_{\pm 1.63}$  &72.84$_{\pm 1.5}$  &66.1$_{\pm 1.9}$ \\
      \cline{3-6}
      &\textbf{VaSCL}&\textbf{78.97}$_{\pm3.69}$   &67.18$_{\pm0.64}$  &\textbf{74.51}$_{\pm0.86}$ &\textbf{68.22}$_{\pm1.71}$\\ \\

      \multirow{3}{*}{\rotatebox[origin=c]{90}{10-Shot}}
      &RoBERTa &85.63$_{\pm2.43}$   &46.55$_{\pm1.84}$  &60.55$_{\pm1.16}$ &57.47$_{\pm0.91}$\\
      &SimCSE& 85.14$_{\pm2.18}$   &72.19$_{\pm0.88}$ &77.13$_{\pm0.76}$ &70.87$_{\pm1.35}$ \\
      \cline{3-6}
      &\textbf{VaSCL} &\textbf{85.34}$_{\pm1.65}$  &\textbf{72.60}$_{\pm0.94}$  &\textbf{78.83}$_{\pm0.51}$ &\textbf{73.70}$_{\pm0.92}$
      \\ \\
      
      \multirow{3}{*}{\rotatebox[origin=c]{90}{20-Shot}}
      &RoBERTa &88.14$_{\pm1.54}$  &51.65$_{\pm1.42}$ &63.51$_{\pm1.08}$ &60.93$_{\pm1.27}$  \\
    %   &{RoBERTa}-W & \\
      &SimCSE& 88.43$_{\pm1.2}$ &75.13$_{\pm0.78}$ &78.59$_{\pm0.78}$ &74.44$_{\pm0.74}$\\
      \cline{3-6}
      &\textbf{VaSCL} &\textbf{89.94}$_{\pm0.89}$ &\textbf{76.60}$_{\pm0.35}$  &\textbf{81.40}$_{\pm0.60}$  &\textbf{77.66}$_{\pm0.64}$
    \end{tabular}}}
    \caption{Few-shot learning  evaluation of Intent Classification. Each result is aggregated over 5 independent splits.  We choose RoBERTa-base as backbone.
    }
    \label{tab:ic_fsl_compare}
  \end{center}
\end{table}

\subsubsection{Evaluation Results}
We report the evaluation results in Tables \ref{tab:STS_compare}, \ref{tab:Cluster_Eval}, and \ref{tab:ic_fsl_compare}. As we can see, both SimCSE and VaSCL largely improve the performance of the pre-trained language models, while VaSCL consistently outperforms SimCSE on most tasks. To be more specific, we attain $0.7\%-2.5\%$ averaged improvement over SimCSE on seven STS tasks, and $1.9\%-4.8\%$ averaged improvement on six short text clustering tasks. We also achieved considerable improvement over SimCSE on intent classification tasks under different few-shot learning scenarios.  We do not include the evaluation on ATIS in Table \ref{tab:ic_fsl_compare} as this dataset is  highly imbalanced with one single dominant class and more than $60\%$ of its classes have than 40 examples. Please refer to Appendix \ref{appendix:full_ic_eval} for more results and details.
% On the other hand, both VaSCL and SimCSE underperform the original pre-trained transformers on ATIS, we attribute this to the fact that  We suspect the performance would reverse when taking the class imbalance into evaluation consideration, which we will report in the updated draft.  

\subsection{Analysis}
To better understand what enables the good performance of VaSCL, we carefully analyze the representations at different semantic granularities.

\begin{figure}[htbp]
    \centering
    \begin{subfigure}{0.49\textwidth}
        \centering
         \includegraphics[scale=0.32]{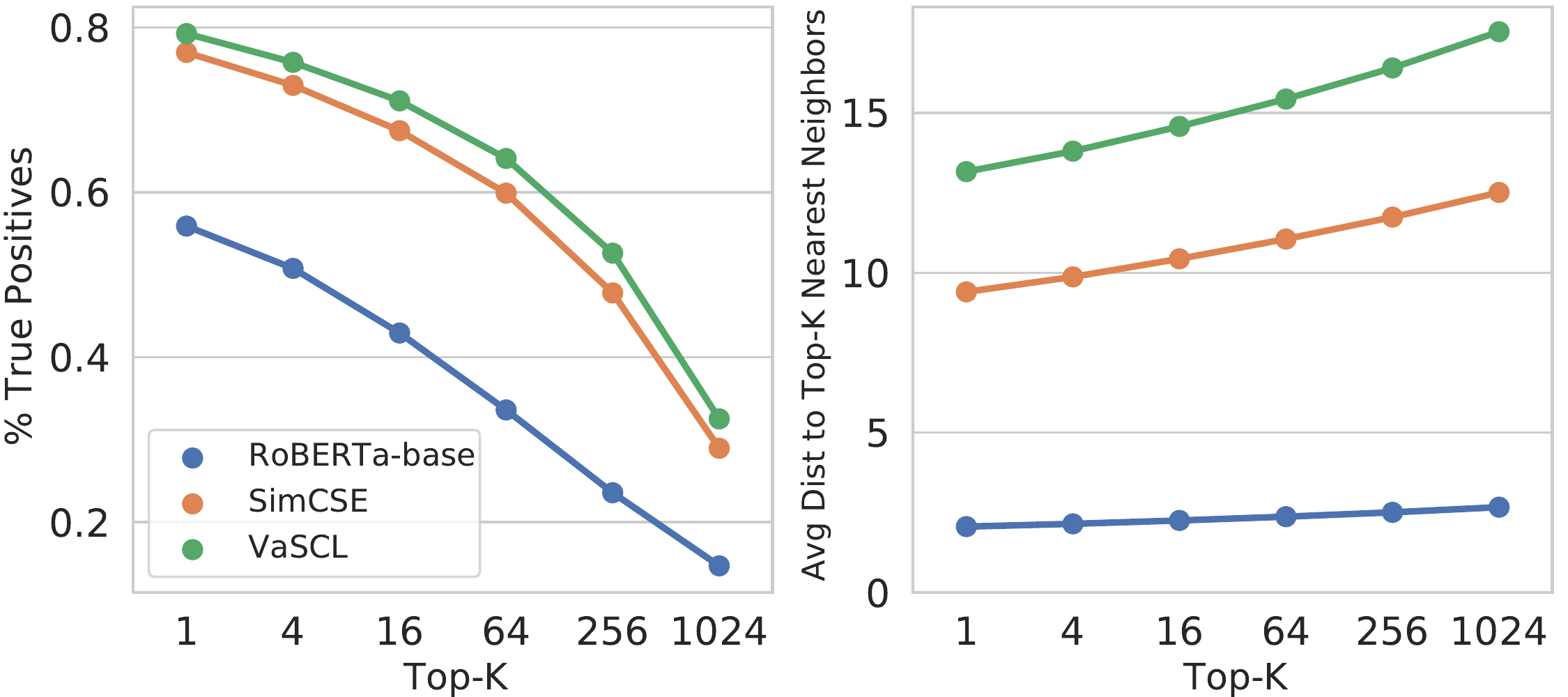}
        \caption{Neighborhood evaluation on StackOverflow. Instances from the same category are treated as true positives.}
        \label{fig:topk-neighbors-metrics}
    \end{subfigure}
    
    \vspace{0.3cm}
    \begin{subfigure}{0.49\textwidth}
        \centering
         \includegraphics[scale=0.32]{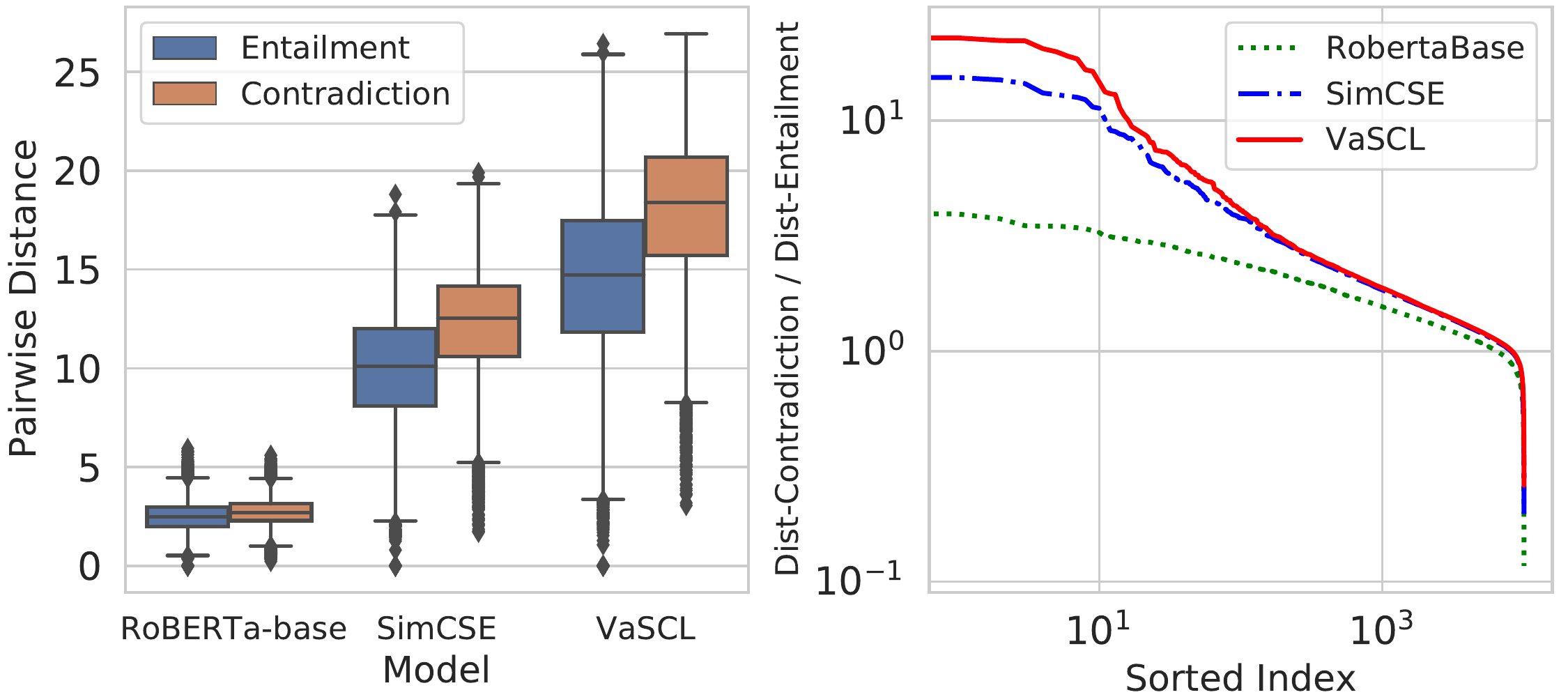}
        \caption{Fine-grained semantics encoding evaluation on NLI.}
        \label{fig:nli-pairwise-metrics}
    \end{subfigure}
    \caption{VaSCL leads to more dispersed representation with data structure being better uncovered.
    }
    \label{fig:embedding_space_analysis}
\end{figure}

\paragraph{Neighborhood Evaluation on Categorical Data}{We first evaluate the neighborhood statistics on StackOverflow \citep{xu2017self} which contains 20 balanced categories, each with 1000 text instances.  For each instance, we retrieve its top-K nearest neighbors in the representation space, among which those from the same class as the instance itself as treated as positives.  In Figure \ref{fig:topk-neighbors-metrics}, we report both the percentage of true positives and the average distance of an instance to its top-K neighbors. For each top-K value, the evaluation is averaged over all 20,000 instances.

As indicated by the small distance values reported in Figure \ref{fig:topk-neighbors-metrics},  the representation space of the original RoBERTa model is more tight and is incapable of uncovering the categorical structure of data. In contrast, both VaSCL and SimCSE are capable of scattering representations apart while better capturing the semantic structures.  Compared with SimCSE, VaSCL leads to even more dispersed representations with categorical structures being better encoded. This is also demonstrated by the better performance attained on both clustering and few-shot learning reported in Tables \ref{tab:Cluster_Eval}\&\ref{tab:ic_fsl_compare}. 
}

\paragraph{Fine-grained Semantic Understanding}{We 
then compare VaSCL against SimCSE and RoBERTa on encoding more fine-grained semantic concepts. We randomly sample 20,000 premises from the combined set of SNLI \citep{bowman2015large} and MNLI \citep{williams2017broad}, where the associated entailment and contradiction hypotheses are also sampled for each premise instance. 
In Figure \ref{fig:nli-pairwise-metrics}, we report both the distribution of the pairwise distances of the entailment or the contradiction pairs \textbf{(left)}. While on the right hand side, we plot the distance of each premise to its entailment hypothesis over that to its contradiction hypothesis \textbf{(right)}. 

We observe the same trend that both SimCSE and VaSCL well separate different instances apart in the representation space, while better discriminating each premise's entailment hypothesis from the contradiction one. Figure \ref{fig:nli-pairwise-metrics} also demonstrates that VaSCL outperforms SimCSE on better capturing the fine-grained semantics when separating different instances apart. This advantage of VaSCL is further validated by Table \ref{tab:STS_compare}, where VaSCL consistently outperforms SimCSE on the STS tasks that require pairwise semantic inference on an even more fine-grained scale. 
}

\begin{figure*}[htbp]
    \centering
    \begin{subfigure}{0.9\textwidth}
        \centering
        \includegraphics[scale=0.30]{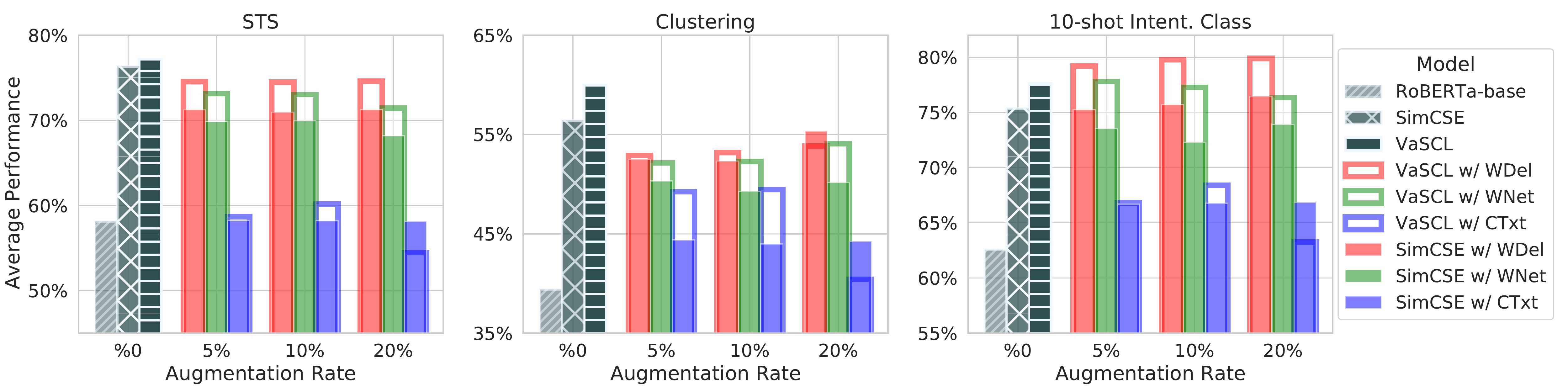}
        \caption{Virtual augmentation vs. explicit augmentation. For each downstream task, we report the mean performance averaged over all its subtasks. The explicit augmentations are evaluated using SimCSE (dropout) for training, \ie "SimCSE w/  \{WDel/WNet/CTxt)\}".}
        \label{fig:expaug_downstream_performance}
    \end{subfigure}
    \vspace{0.2cm}
    
    \begin{subfigure}{0.9\textwidth}
        \centering
         \includegraphics[scale=0.28]{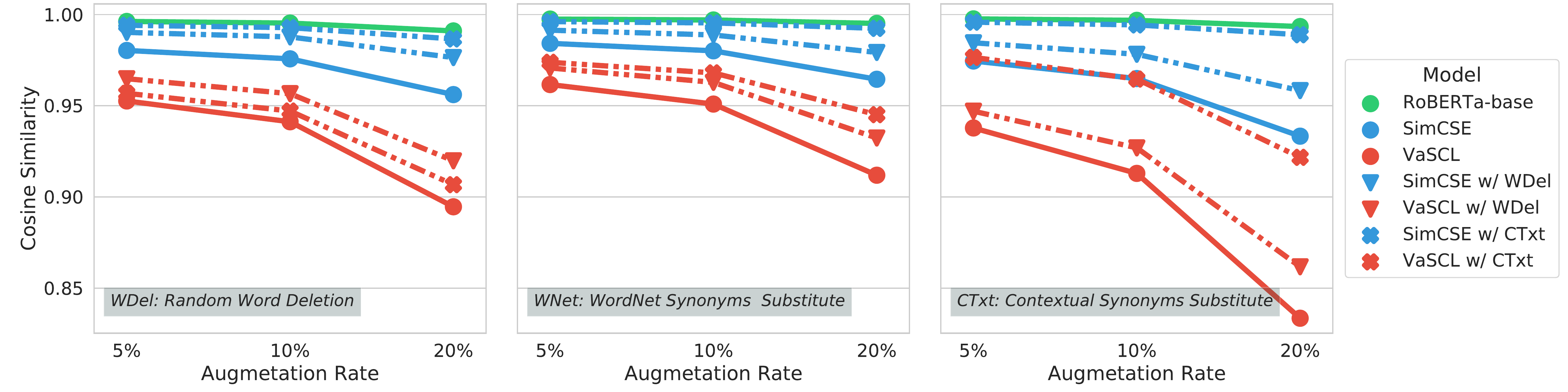}
        \caption{Cosine similarity between the each original training example and its augmentation evaluated on the representation spaces of different models. From left to right, the augmentations are obtained via \emph{WDel}, \emph{WNet}, and \emph{CTxt}. Each point is averaged over 20,000 randomly sampled training examples and the associated augmentations. We exclude  \emph{"SimCSE w/ WNet"} and \emph{"VaSCL w/ WNet"} for better visualization. Please refer to Appendix \ref{appendix:expaug_full_plot} for the full plot. }
        \label{fig:expaug_cossim}
    \end{subfigure}
    \caption{Comparing and combining virtual augmentation with explicit augmentation.
    }
    \label{fig:expaug_eval}
\end{figure*}

\subsection{Explicit Data Augmentation}
\label{subsec:expaug}
% Data augmentation plays a key role in contrastive representation learning, and therefore it raises a question when applying them to NLP where no general rules are established for effective data augmentation due to the discrete the nature of natural languages. 
To better evaluate our virtual augmentation oriented VaSCL model, we compare it against different explicit data augmentation strategies that directly operate on the discrete text. 
% Our focus is twofold: to test whether VaSCL is more effective than the common yet popular explicit augmentation strategies; and to investigate whether the composition of virtual and explicit augmentations would lead to better sentence representations.
Specifically, we consider the following approaches:\footnote{They are implemented using the nlpaug library \url{https://github.com/makcedward/nlpaug}.} \underline{\emph{WDel}} (random word deletion) removes words from the input text randomly;   \underline{\emph{WNet}} (WordNet synonym substitute) transforms a text instance by replacing its words with the WordNet synonyms
\citep{morris2020textattack,ren2019generating}; and \underline{\emph{CTxt}} (contextual synonyms substitute) leverages the pre-trained transformers to find top-n suitable words of the input text for substitution  \citep{kobayashi2018contextual}. 
For each strategy, we evaluate three augmentation strengths by partially changing $5\%$, $10\%$, and $20\%$ words of each text instance. For a positive pair $(x_i, x_i')$, $x_i$ denotes the original text and $x_{i'}$ is the augmentation.
%  In this setting, the original VaSCL and SimCSE are indicated with augmentation strength equal $0\%$, \ie $x_i$ and $x_i'$ are the same.
\dejiao{We also explored the case where both $x_i$ and $x_{i'}$ are the transformations of the original text, which we find yielding worse performance.}

\paragraph{Virtual Augmentation Performs Better}{The performance of explicit text augmentation is evaluated using the standard dropout for training, \ie "SimCSE w/  \{WDel/WNet/CTxt)\}" in Figure \ref{fig:expaug_eval}. As Figure \ref{fig:expaug_downstream_performance} shows, contrastive learning with moderate explicit text augmentations, \ie augmentation strength less than $20\%$, does yield better sentence representations when compared with the original RoBERTa model. Nevertheless, both virtual augmentation strategies, \ie SimCSE \& VaSCL, substantially outperform all three explicit text augmentation strategies on almost all downstream tasks. Although a bit surprising, especially considering the performance gap between SimCSE and explicit augmentations, this comparison provides a different perspective on interpreting the underlying challenge of designing effective transformations that operate on the discrete text directly.} 

\paragraph{VaSCL Outperforms SimCSE}{Figure \ref{fig:expaug_downstream_performance} also empirically demonstrates that VaSCL outperfoms SimCSE no matter in presence of explicit text augmentations or not. The only exceptions occur when the explicit augmentation strength is too large, \ie $20\%$ of the words of each text are perturbed. One possible explanation is that undesired noises are generated by large perturbations on discrete texts directly, which can violate the coherent semantics maintained by a  neighborhood and hence make it hard for VaSCL to generate effective virtual augmentations.}

\paragraph{New Linguistic Patterns Are Required For A Win-Win Performance Gain}{Another observation draw from Figure \ref{fig:expaug_downstream_performance} is that both SimCSE and VaSCL attain worse performance on most downstream tasks when combining with explicit text augmentations. Although VaSCL does improve the performance of explicit augmentations in most cases, this is undesired as we expect a win-win outcome that moderate explicit augmentations could also bolster the performance of VaSCL. We hypothesize that new yet informative linguistic patterns are missing for the expected performance gain. 

To validate our hypothesis, in Figure \ref{fig:expaug_cossim} we report the cosine similarity between each original training example and its augmentation evaluated on the representation spaces of different models. Our observation is twofold. Frist, the representations induced by RoBERTa and the one trained with contextual synonyms substitution ("SimCSE w/ CTxt") are very similar in all three settings, which also explains why "SimCSE w/ WDel" attains similar performance as RoBERTa on the downstream tasks.  We attribute this to the fact that CTxt leverages the transformer itself to generate augmentations which hence carry limited unseen and effective linguistic patterns.  Second, as indicated by the comparatively smaller similarity values in Figure \ref{fig:expaug_cossim}, the incorporation of explicit augmentations tightens the representation spaces of both SimCSE and VaSCL, which also in turn results in worse performance of downstream tasks. One possible explanation is that all three explicit augmentations are weak and noisy, which harm both the instance discrimination force  and the semantic relevance of each neighborhood. 
}

% \citet{longpre-etal-2020-effective} show that many data augmentation methods are not able to achieve gains when using large pre-trained language models, as they are already invariant to various transformations by themselves. They hypothesize that data augmentation methods can only be really beneficial if they are creating new linguistic patterns that have not been seen before.

\section{Conclusion}
In this paper, we present an virtual augmentation oriented contrastive learning framework for unsupervised sentence representation learning.  Our key insight is that data augmentation can be interpreted as constructing the neighborhoods of each training instance, which can in turn be leveraged to generate effective data augmentations. We evaluate our VaSCL model on a wide range of downstream tasks and substantially advance the state-of-the-art results. Moreover, we conduct in-depth analysis and show that VaSCL leads to a more dispersed representation space with the data semantics at different granularities being better encoded. 

On the other hand, we observe performance drop of both SimCSE and VaSCL when incorporating explicit text augmentations into the training set. We suspect this is caused by the linguistic patterns generated by explicit augmentations are less diverse, yet noisy. We hypothesize effective data augmentation operations on the discrete texts  could complement our virtual augmentation approach if new and informative linguistic patterns are generated. 

% when evaluated on various downstream tasks that involve understanding semantic concepts at different granularities.

\bibliography{anthology,custom}
\bibliographystyle{acl_natbib}

\clearpage 
\newpage 

\appendix

\section{Implementation}
\label{appendix:implementation}
Same as the original SimCSE work \citep{gao2021simcse}, we adopted
$10^6$ randomly sampled sentences from English
Wikipedia as training data.\footnote{We download the training data via \url{https://github.com/princeton-nlp/SimCSE/blob/main/data/download_wiki.sh}.} 
We implement our models with Pytorch \citep{paszke2017automatic}. We use the pre-trained RoBERTa models as the backbone. We choose a two-layer MLP with size ($d\times d$, $d\times 128$) to optimize our the contrastive learning losses, where $d$ denotes the dimension of the sentence representations. We use Adam \citep{KingmaB14} as our optimizer with a constant learning rate of $5e$-04 which we scale it $5e$-06 for updating the backbones/transformers. We set the virtual augmentation strength of VaSCL, \ie $\Delta$ in Equation (\ref{eq:optimal_delta}),   to 15 for both DistilRoBERTa and RoBERTaBase,  and 30 for RoBERTaLarge. 

We train SimCSE \citep{gao2021simcse} using $3e$-05 for optimizing the contrastive learning head and the backone. We also tried the default learning rate $1e$-05 (suggested in \citet{gao2021simcse}) as well as our learning rate setup for optimizing the RoBERTa models with SimCSE. We found $3e$-05 yield better performance. For both SimCSE and VaSCL, we set the batch size to 1024, and train all models over 5 epochs and evaluate on the development set of STS-B every 500 iterations. We report all our evaluations on the downstream tasks with the associated checkpoints attaining the best performance on validation set of STS-B.

\section{Dataset Statistics}
\label{appendix:data_statistics}
\subsection{Intent Classification Dataset}
\label{appendix:IC_datastats}
% \begin{table*}[!htbp]
%   \begin{center}
%     \begin{tabular}{l|cccccc}
%     \hline
%     Dataset & Train &Dev &Test & D  & C & ImN  \\
%     \hline
%     ATIS  & \\
%     SNIPS &13084 &700 & 700& 1 &7 &1   \\
%     HWU64 &8954 & 1076 & 1076 &21 &64 &  \\
%     CLINC150 & 15000& 3000 &4500 & 10  &150 & 1    \\
%     BANKING77 & 8622 & 1540 &3080 &  & 77&  \\
%       \hline
%     \end{tabular}
%     \caption{Statistics of five intent classification datasets. D: number of domains; C: number of classes; ImN: imbalance number defined as the size of the largest class divided by that of the smallest class.}
%     \label{tab:intent_datastats}
%   \end{center}
% \end{table*}

\begin{figure*}[htbp]
    \centering
    \begin{subfigure}{0.9\textwidth}
        \centering
        \includegraphics[scale=0.30]{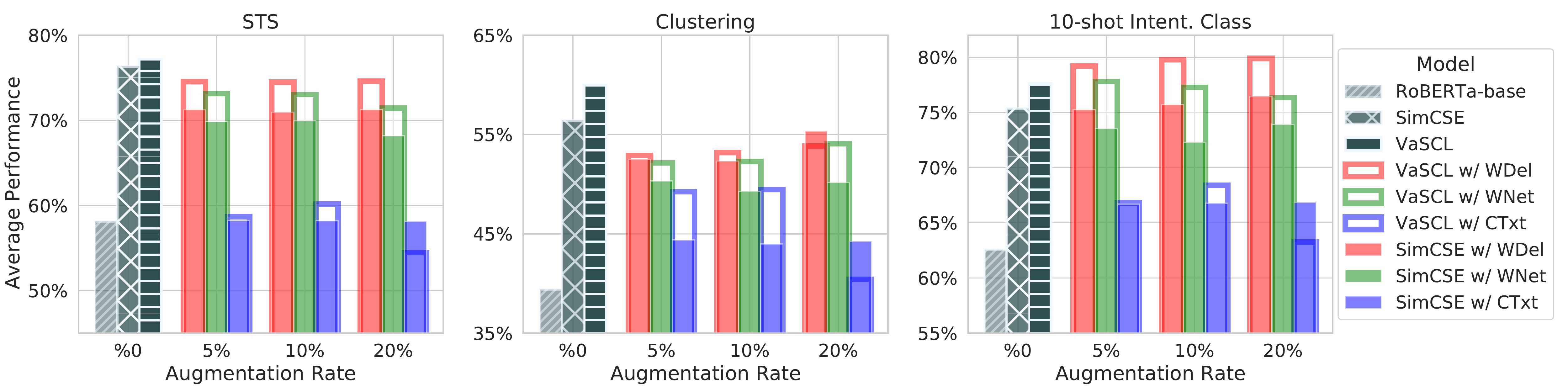}
        \caption{Evaluating VaSCL in presence of different explicit data augmentation strategies.}
        \label{fig:expaug_downstream_performance_appendix}
    \end{subfigure}
    \vspace{0.2cm}
    
    \begin{subfigure}{0.9\textwidth}
        \centering
         \includegraphics[scale=0.28]{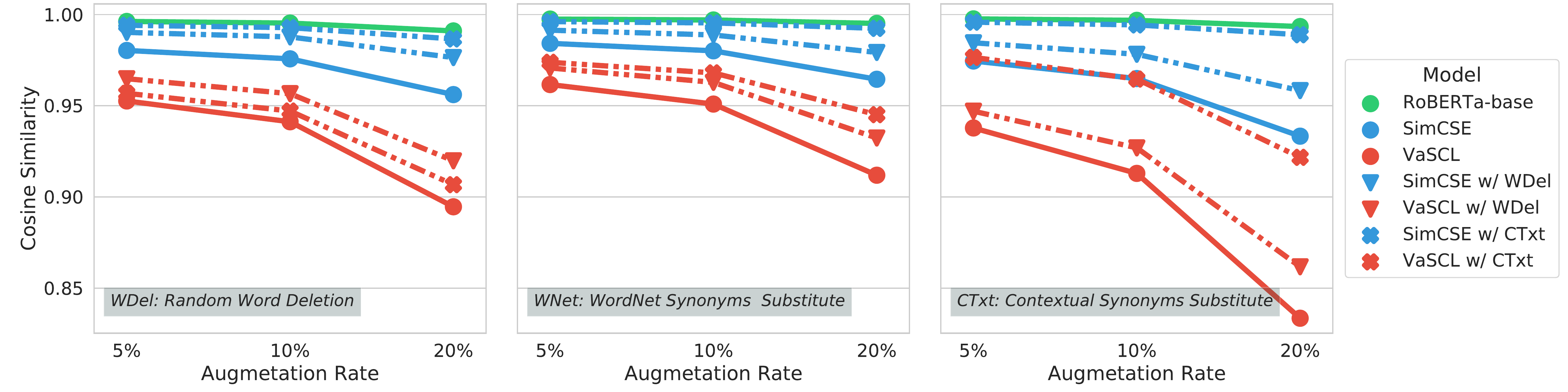}
        \caption{Cosine similarity between the representations of each original training example and its augmentation evaluated on different models. From left to right, the augmentations are obtained via \emph{WDel}, \emph{WNet}, and \emph{CTxt}. Each point is averaged over 20,000 randomly sampled training examples.  }
        \label{fig:expaug_cossim_full}
    \end{subfigure}
    \caption{Comparing and combining virtual augmentation with explicit text augmentations. (Full plot of Figure \ref{fig:expaug_eval} in Section \ref{subsec:expaug}.)
    }
    \label{fig:expaug_eval_aug}
\end{figure*}

We evaluate our model on four intent classification datasets: (1) \textbf{SNIPS} \citep{coucke2018snips} is a SLU benchmark that consists of 7 distinct intents. (2) \textbf{BANKING77} \citep{casanueva2020efficient}  is a large fine-grained single banking domain intent dataset with 77 intent classes.
(3) \textbf{HWU64} \citep{liu2021benchmarking} contains 25,716 examples for 64 intents in 21 domains. (4) \textbf{CLINC150} \citep{larson2019evaluation} spans 150 intents and 23,700 examples across 10 domains. As we can see here, SNIPS are limited to only a small number of classes, which oversimplifies the intent detection task and does not emulate the true environment of commercial systems. In comparison, the remaining three datasets contain much more diversity and are more challenging.

\subsection{Short Text Clustering Dataset}
\label{appendix:clustering_datastats}
\begin{table}[!htbp]
  \begin{center}
    {\small{\begin{tabular}{l|cccc}
    \hline
    Dataset & N & $\bar{W}$  & C & ImN  \\
    \hline
    AgNews  &8.0K &23 &4 & 1 \\
    SearchSnippets & 12.3K &18 & 8 & 7  \\
    StackOverflow & 20K &8 & 20 &1    \\
    Biomedical & 20K &13 &  20 & 1 \\
    GoogleNews  & 11.1K & 28 & 152 & 143   \\
    Tweet & 2.5K & 8 & 89 & 249 \\
      \hline
    \end{tabular}}}
    \caption{Statistics of six short text clustering datasets. N: number of text samples; $\bar{W}$: average number of words each text example has; C: number of clusters; ImN: imbalance number defined as the size of the largest class divided by that of the smallest class.}
    \label{tab:cluster_datastats}
  \end{center}
\end{table}

\begin{itemize}
    \item \textbf{SearchSnippets} is extracted from web search snippets, which contains 12340 snippets associated with 8 groups \citet{phan2008learning}.
    
    \item \textbf{StackOverflow} is a subset of the challenge data published by Kaggle\footnote{https://www.kaggle.com/c/predict-closed-questions-on-stackoverflow/download/train.zip}, where 20000 question titles associated with 20 different categories are selected by \citet{xu2017self}.

    \item \textbf{Biomedical} is a subset of PubMed data distributed by BioASQ\footnote{http://participants-area.bioasq.org}, where 20000 paper titles from 20 groups are randomly selected by \citet{xu2017self}.

    \item \textbf{AgNews} is a subset of news
titles \citep{zhang2015text}, which contains 4 topics selected by \citet{rakib2020enhancement}.

    \item \textbf{Tweet} consists of 89 categories with 2472 tweets in total \citep{yin2016model}.

    \item \textbf{GoogleNews} contains titles and snippets of 11109 news articles related to 152 events \citep{yin2016model}. 
\end{itemize}

\section{Full Evaluation of Intent Classification}
\label{appendix:full_ic_eval}

\textbf{ATIS} \citep{hemphill1990atis} is a benchmark for the air travel domain. This dataset is highly imbalanced, with the largest class contains $73\%$ of all the training and validation examples. Moreover, there are more than $60\%$ classes have less than 20 examples. In order to conduct the evaluation reported in Table \ref{tab:ic_fsl_compare_full}, we truncate those categories of the merged training and validation set with less than 40 examples, which is the minimum requirement for Table \ref{tab:ic_fsl_compare_full}. Consequently, there are only eight classes remained.   

As shown in Table \ref{tab:ic_fsl_compare_full}, both SimCSE and VaSCL largely improve the performance of the intent classification tasks under different few-shot learning scenarios. However, they also underperform the original pre-trained transformers on ATIS. We attribute this to the highly imbalanced classes distribution of ATIS. We suspect the performance would reverse when taking the class imbalance into the evaluation consideration, which we will report in the updated draft.

\begin{table*}[htbp]
  \begin{center}
    \begin{tabular}{llcccccc}
    %   & \multicolumn{2}{c}{\textbf{SNIPS}}& \multicolumn{2}{c}{\textbf{ATIS}}& \multicolumn{2}{c}{\textbf{BANKING77}}&\multicolumn{2}{c}{\textbf{CLINC150}}& \multicolumn{2}{c}{\textbf{HWU64}} \\
    & &\textbf{SNIPS}&\textbf{ATIS} & \textbf{BANK77} &\textbf{CLINC150} & \textbf{HWU64} \\
      \cline{3-7}
      \multirow{3}{*}{\rotatebox[origin=c]{90}{5-Shot}}
    %   &GloVe & \\ 
      &RoBERTa &76.71$\pm$4.84 &\textbf{45.42}  $\pm$12.42  &38.77$\pm$ 2.29  &55.19  $\pm$1.99 &51.52  $\pm$2\\
    %   &{RoBERTa}-W & \\
      &SimCSE&76.94  $\pm$2.53 &36.75$\pm$17.03  &\textbf{67.48}  $\pm$1.63  &72.84 $\pm$1.5  &66.1$\pm$1.9 \\
      \cline{3-7}
      &\textbf{VaSCL}&\textbf{78.97}$\pm$3.69 &35.78$\pm$12.65  &67.18$\pm$0.64  &\textbf{74.51}$\pm$0.86  &\textbf{68.22}$\pm$1.71\\ \\

      \multirow{3}{*}{\rotatebox[origin=c]{90}{10-Shot}}
      &RoBERTa &85.63$\pm$2.43  &\textbf{58.75} $\pm$9.35 &46.55  $\pm$1.84  &60.55 $\pm$1.16 &57.47$\pm$0.91\\
    %   &{RoBERTa}-W & \\
      &SimCSE& 85.14$\pm$2.18 &44.95$\pm$14.93  &72.19$\pm$0.88 &77.13$\pm$0.76 &70.87$\pm$1.35 \\
      \cline{3-7}
      &\textbf{VaSCL} &\textbf{85.34} $\pm$1.65 &46.80$\pm$13.11 &\textbf{72.60}$\pm$0.94  &\textbf{78.83}$\pm$0.51 &\textbf{73.70}$\pm$0.92
      \\ \\
      
      \multirow{3}{*}{\rotatebox[origin=c]{90}{20-Shot}}
      &RoBERTa &88.14$\pm$1.54  &\textbf{72.15}$\pm$7.55  &51.65$\pm$1.42 &63.51$\pm$1.08 &60.93$\pm$1.27  \\
    %   &{RoBERTa}-W & \\
      &SimCSE& 88.43$\pm$1.2  &61.45$\pm$8.83 &75.13$\pm$0.78 &78.59$\pm$0.78 &74.44$\pm$0.74\\
      \cline{3-7}
      &\textbf{VaSCL} &\textbf{89.94} $\pm$0.89 &62.12$\pm$7.09 &\textbf{76.60}$\pm$0.35  &\textbf{81.40}$\pm$0.60  &\textbf{77.66}$\pm$0.64
    \end{tabular}
    \caption{Few-shot learning  evaluation of Intent Classification. We choose RoBERTaBase as backbone.
    % $\diamondsuit$ and $\spadesuit$: results evaluated on the checkpoints provided by \citep{reimers-2019-sentence-bert} and \citep{gao2021simcse}, respectively. 
    }
    \label{tab:ic_fsl_compare_full}
  \end{center}
\end{table*}

\section{Comparing and Combining with Explicit Augmentations}
\label{appendix:expaug_full_plot}
Please refer to Figure \ref{fig:expaug_cossim_full} to the completed plot of Figure \ref{fig:expaug_cossim} in Section \ref{subsec:expaug}.

\end{document}